%% file: main.tex
\documentclass{article}

\usepackage{microtype}
\usepackage{graphicx}
\usepackage{booktabs} 
\usepackage{caption}
\usepackage{subcaption}
\usepackage{enumitem}
\usepackage{arydshln}
\usepackage{amsmath}
\usepackage{fixltx2e}
\usepackage{multirow}
\usepackage{bm}

\newcommand{\figref}[1]{Fig.~\ref{#1}}
\newcommand{\tabref}[1]{Tab.~\ref{#1}}
\newcommand{\secref}[1]{Sec.~\ref{#1}}

\usepackage{hyperref}

\usepackage[accepted]{icml2021}

\icmltitlerunning{A Tale of Two Resampling Strategies for Long-Tailed Detection}

\begin{document}

\twocolumn[
\icmltitle{Image-Level or Object-Level?\\A Tale of Two Resampling Strategies for Long-Tailed Detection}


\icmlsetsymbol{intern}{*}

\begin{icmlauthorlist}
\icmlauthor{Nadine Chang}{cmu,intern}
\icmlauthor{Zhiding Yu}{nv}
\icmlauthor{Yu-Xiong Wang}{uiuc}
\icmlauthor{Anima Anandkumar}{nv,caltech}
\icmlauthor{Sanja Fidler}{nv,toronto,vector}
\icmlauthor{Jose M. Alvarez}{nv}
\end{icmlauthorlist}

\icmlaffiliation{cmu}{Carnegie Mellon Univ}
\icmlaffiliation{nv}{NVIDIA}
\icmlaffiliation{uiuc}{Univ of Illinois at Urbana-Champaign}
\icmlaffiliation{caltech}{Caltech}
\icmlaffiliation{toronto}{Univ of Toronto}
\icmlaffiliation{vector}{Vector Institute}

\icmlcorrespondingauthor{Zhiding Yu}{zhidingy@nvidia.com}
\icmlcorrespondingauthor{Nadine Chang}{nadinec@cmu.edu}


\vskip 0.3in
]



\printAffiliationsAndNotice{\icmlIntern} 

\begin{abstract}
Training on datasets with long-tailed distributions has been challenging for major recognition tasks such as classification and detection. To deal with this challenge, image resampling is typically introduced as a simple but effective approach. However, we observe that long-tailed detection differs from classification since multiple classes may be present in one image. As a result, image resampling alone is not enough to yield a sufficiently balanced distribution at the object level. We address object-level resampling by introducing an object-centric memory replay strategy based on dynamic, episodic memory banks. Our proposed strategy has two benefits: 1) convenient object-level resampling without significant extra computation, and 2) implicit feature-level augmentation from model updates. We show that image-level and object-level resamplings are both important, and thus unify them with a joint resampling strategy (RIO). Our method outperforms state-of-the-art long-tailed detection and segmentation methods on LVIS v0.5 across various backbones. Code is available at \url{https://github.com/NVlabs/RIO}.
\end{abstract}

\input{sec1_intro}
\input{sec2_related}
\input{sec3_method}
\input{sec4_exper}
\input{sec5_concl}
\input{sec6_ack}

\bibliography{ref}
\bibliographystyle{icml2021}

\end{document}

%% file: sec1_intro.tex
\section{Introduction}

Real-world visual data often follows a long-tailed distribution, where a few object classes are very common and many of the classes are rare~\cite{zhu2014capturing,liu2019large,gupta2019lvis}. However, many existing datasets are curated to be balanced~\cite{krizhevsky2009learning,pascal-voc-2012,lin2014microsoft}, leading to a discrepancy between the performance of the methods developed and tested on these datasets and their performance when deployed in the real world. In many applications, localization/recognition of rare classes is critical, e.g., in autonomous driving the ego-car is expected to be able to localize and react to objects such as certain animals that the system has not often seen in the training data. In this paper, we address the problem of long-tailed object detection (and segmentation).

\begin{figure}[t]
\centering
\includegraphics[width=\linewidth]{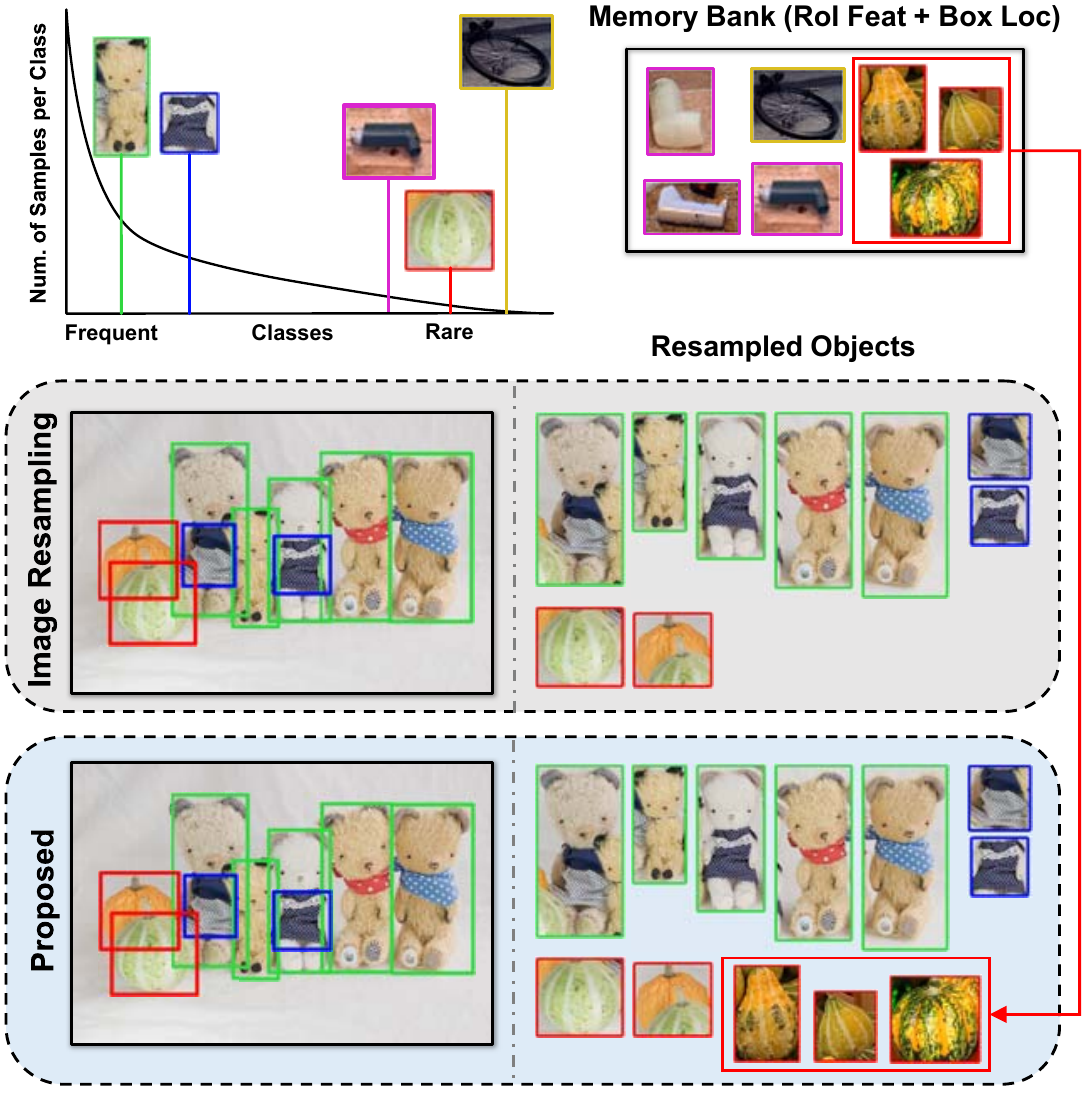}
\caption{We illustrate the difference between image resampling and our proposed resampling strategy based on object-centric memory replay. In image resampling, all the objects within a repeated image are resampled, including the teddy bears, aprons, and gourds in this example. Our method increases the sampling of the rare gourds by retrieving additional ones from the memory bank.}
\label{fig:teaser}
\end{figure}

Long-tailed recognition has been recently popularized, facilitated by the release of large-scale visual recognition benchmarks~\cite{liu2019large,gupta2019lvis}. Popular approaches include reweighting the loss function according to the class frequency~\cite{tan2020equalization}, decoupling representation and classifier~\cite{kang2019decoupling}, as well as few-shot learning techniques such as feature normalization~\cite{chen2019closer,wang2020frustratingly}, fine-tuning~\cite{wang2020frustratingly}, and meta-learning~\cite{wang2017learning}. In addition to these techniques, repeat factor sampling (RFS)~\cite{mahajan2018exploring, gupta2019lvis} has emerged as a simple but popular image resampling baseline. By repeating images containing rare classes in each epoch, the method is shown to work well on long-tailed classification problems~\cite{wang2018low, kang2019decoupling}.
However, applying the same method out of the box for object detection, even though shown to be empirically effective~\cite{gupta2019lvis}, ignores the fact that a general image often contains multiple objects from different classes. This is aggravated by the common co-occurrence of frequent classes and rare classes, as shown in~\figref{fig:teaser}. Therefore, image resampling alone is not sufficient to successfully balance the training distribution.

One special aspect of detection is that objects are the basic units. This requires object-centric sampling strategies that can balance the distribution at the object-level. Achieving this goal is not straightforward, because image-level batch sampling is still the predominant pipeline for the current object detection training practices. Arguably, one could consider straightforward modifications that are compatible to this pipeline, for example, sampling additional images or even object crops containing the target rare classes. However, this may introduce additional computational costs by the extra forward and backward propagation during training, especially when the distribution is highly imbalanced. Other possible complications include the need to ignore frequent classes when taking the whole sampled image as input, or losing image context when just taking the object crops.

We address the issues by proposing a novel object-centric memory replay framework that augments each batch with additional region of interest (RoI) features and their corresponding box locations. At the core of our framework is the idea of reusing the RoI features from the previous forward propagation. This can be achieved by introducing a dynamic memory bank that stores the RoI features and box coordinates. The memory bank is continuously updated every iteration by pushing with the batch of newly computed RoI features, and basically serves as a buffer with limited size to accumulate the RoI features for every class. Under an imbalanced input distribution, the buffers of frequent classes get filled and updated quicker, whereas the ones of rare classes also get filled over time (but slower). Thus the memory bank allows for efficient object resampling without requiring additional forward/backward propagation. The above strategy has two additional advantages: (1) The memory bank contains features from historical model snapshots, thus achieving model-level augmentation across time.  (2) The box locations for the same object can be slightly different over time, because models are continuously updated and images are randomly augmented. Thus our OCS framework can achieve diversified augmentation at various levels.

We provide detailed analysis on both image-level and object-level resamplings to examine the overlooked pitfalls for detection. Although the two strategies capture different aspects of resampling, they are actually not mutually exclusive. Although a memory bank can counter imbalanced distribution by accumulating features over time, our analysis indicates that certain extremely rare classes only appear few times in one epoch. This leads to significantly outdated features and decreases the memory bank quality. Fortunately, such issue can be directly addressed with an RFS-based image resampling. We therefore argue that image-level and object-level resamplings can be symbiotic, which leads to a joint resampling strategy termed \textbf{RIO} (Resampling at Image-level and Object-level, shown in \figref{fig:teaser}). 

We showcase our method's efficacy on LVIS which is currently the most popular and challenging benchmark for long-tailed detection~\citep{gupta2019lvis}. Previous methods improve the overall accuracy but the rare classes still show significant gaps to the common and frequent classes. Our method greatly alleviates this issue, leading to state-of-the-art performance both overall and on rare classes.

\textbf{Summary of Contributions:}
\vspace{-0.1in}
\begin{itemize}[leftmargin=*]
    \item We propose a novel object-centric memory replay strategy based on memory bank. Our method is able to perform efficient object-level sampling with implicit augmentation.
    \item We show the importance of balancing the distribution at both image-level and object-level in object-centric memory replay. This motivates us to propose RIO, a unified resampling framework comprising both schemes.
    \item Our framework is frustratingly simple but motivated. The proposed method achieves state-of-art performance in both the overall accuracy and accuracy of rare classes.
\end{itemize}

%% file: sec2_related.tex
\section{Related Work}

\textbf{Few-shot Learning.}
Few-shot learning has been a dominant task towards addressing imbalanced datasets. Many approaches focus on utilizing similar class features to assist in creating better rare class features or scores~\cite{vinyals2016matching, snell2017prototypical}. More recent approaches attempt to address few-shot learning as a ``learning to learn" task using meta-learning~\cite{finn2017model,wang2016learning, zhang2018metagan}. Finally, another stream attempts to solve few-shot learning by adding more training samples. This is similar in nature to resampling, but explicitly attempts to create samples that are more unique. Some works use Generative Adversarial Networks to synthesize entirely new images~\cite{radford2015unsupervised}, while others require additional annotations to create domain specific samples~\cite{lake2013one, radford2015unsupervised}. Some work bypasses the complexity of generating images and instead uses meta-learning to hallucinate features~\cite{wang2018low}. Similarly, our method also focuses on feature-level augmentation. While few-shot learning has seen significant advance, it assumes that imbalanced datasets contain either large or small amounts of data and differs from long-tailed learning.

\textbf{Long-tailed Classification.}
Certain limitations of few-shot learning have motivated the interest to study long-tailed learning. A popular approach to address long-tailed learning is to transfer feature representations~\cite{liu2019large} or intra-class variance~\cite{yin2019feature} from common categories to rare ones. However, these approaches tend to require relatively complex frameworks such as modulated attention and dynamic meta-embeddings~\cite{liu2019large}. Similar frameworks are also used to address the additional intra-class variance in long-tailed datasets and improve the robustness and generalization for the rare categories ~\cite{zhu2020inflated}. Recent works indicate that simple strategies such as resampling and reweighting in different stages of training also work very well~\cite{kang2019decoupling,shen2016relay,cao2019learning}. \cite{kang2019decoupling} shows that normalizing and adjusting classifiers only with \textbf{resampling} strategies are able to achieve good performance. In addition, the class-balance loss shows that class-balanced training can be achieved by re-weighting category losses based on the category size~\cite{cui2019class}. Both resampling and rebalancing aim towards a common goal of balancing the contribution of all classes. Their overall success indicates that class balancing is crucial for long-tailed learning.

\textbf{Long-tailed Detection and Segmentation.}
Long-tailed detection and segmentation recently raised great attention. A popular long-tailed detection baseline is Repeat Factor Sampling (RFS) which repeats an image based on the rarest object within that image. The class-aware resampling method attempts to balance the batch by filling it with classes first, and then randomly sampling images for each class~\cite{shen2016relay}. However, class-aware sampling often performs worse when there is a large number of classes with a highly imbalanced sample distribution, such as LVIS. Classic fine-tuning~\cite{wang2020frustratingly}, a baseline also for few-shot learning, has shown that simply fine-tuning a pretrained model on the rarer categories substantially increases their performance. Fine-tuning also uses RFS to fine-tune on more rare samples. Additionally, Hu et al. separates the long-tailed dataset into sections and performs incremental learning in each section~\cite{hu2020learning}. Another major line of work focuses on different loss strategies.
Equalization Loss (EQL)~\cite{tan2020equalization} and Seesaw Loss~\cite{wang2020seesaw} attempt to balance encouraging and discouraging gradients for classes. Similar in spirit, another method proposes a balanced group softmax (BAGS) and tries to balance all classifiers such that both frequent and rare classifiers are trained sufficiently~\cite{li2020overcoming}. Lastly, Forest R-CNN~\cite{wu2020forest} leverages the existing hierarchical structure in the dataset and constructs a different model architecture based on the class hierarchies. Although performing well on LVIS, the proposed method is highly reliant on the extra supervision from the hierarchy which in reality is not always available or requires careful construction.

\textbf{Memory Bank.}
Memory bank was recently proposed by Wu et al.~\cite{wu2018unsupervised}. A memory bank consists of the feature representations of dataset samples, supporting fast large-scale feature retrievals without forward and back-propagation computation costs. Though widely used in recent unsupervised contrastive representation learning methods~\cite{chen2020simple,he2020momentum,chen2020improved}, the technique has not been applied to long-tailed object detection problems to the best of our knowledge.

%% file: sec3_method.tex
\section{Method}

\begin{figure*}[t]
\centering
\includegraphics[width=0.95\linewidth]{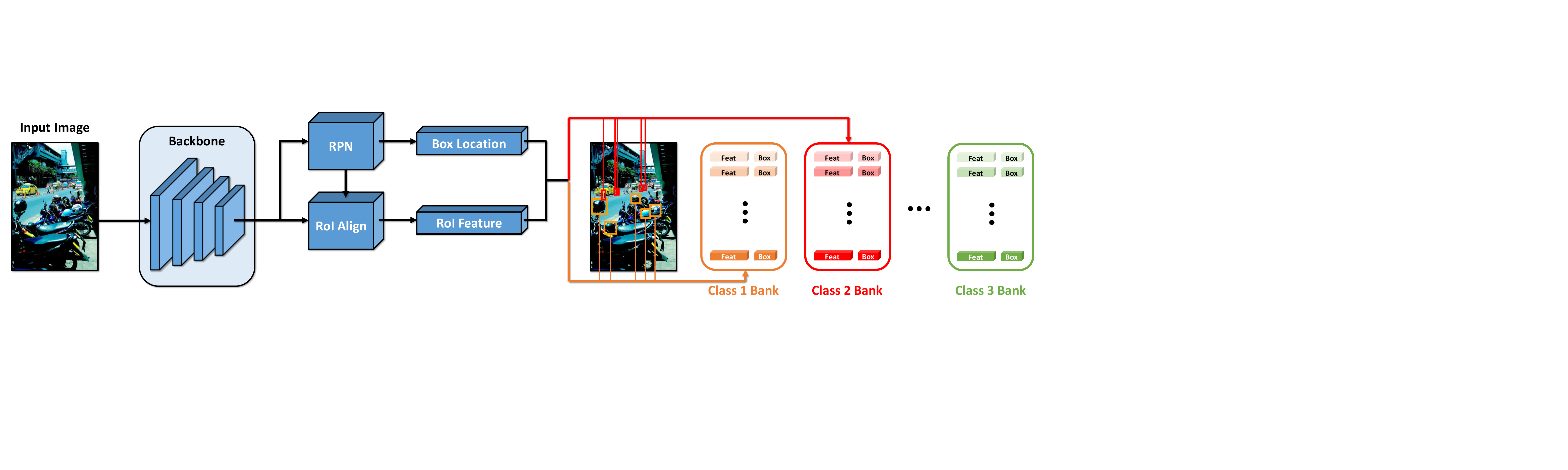}
\caption{A depiction of RoI object feature's extraction location and our memory bank \textit{M} used for object resampling.}
\label{fig:bank}
\end{figure*}

Our ultimate goal is to boost the number of rare classes only, which cannot be achieved by using only image resampling due to the nature of multi-object images. Thus, in junction with image sampling, batches are augmented with RoI object samples from the memory bank as seen in~\figref{fig:bank}. We describe our memory bank setup, dynamic changes, and training in the following section.

Following LVIS class set definitions, the classes of a long-tailed dataset can be divided into three sets of classes based on the number of training examples per class: 1) $\bm{S}_f$, frequent classes with $>100$, 2) $\bm{S}_c$, common classes with $< 100$ images but $> 10$ images, and 3) $\bm{S}_r$, rare classes with $\leq 10$ images. We denote the total number of images in the dataset as $N$. For a single image $I_i$, $i \in (1, \dots ,N), $we denote all \textit{k} objects as $o^j_i$, $j \in (1, \dots ,k)$. Each $o^j_i$ corresponds to its category $c^j_i$, $j \in (1, \dots ,k)$.

\subsection{Memory Bank} \label{sec:bank}
\begin{figure}[t]
\centering
\includegraphics[width=\linewidth]{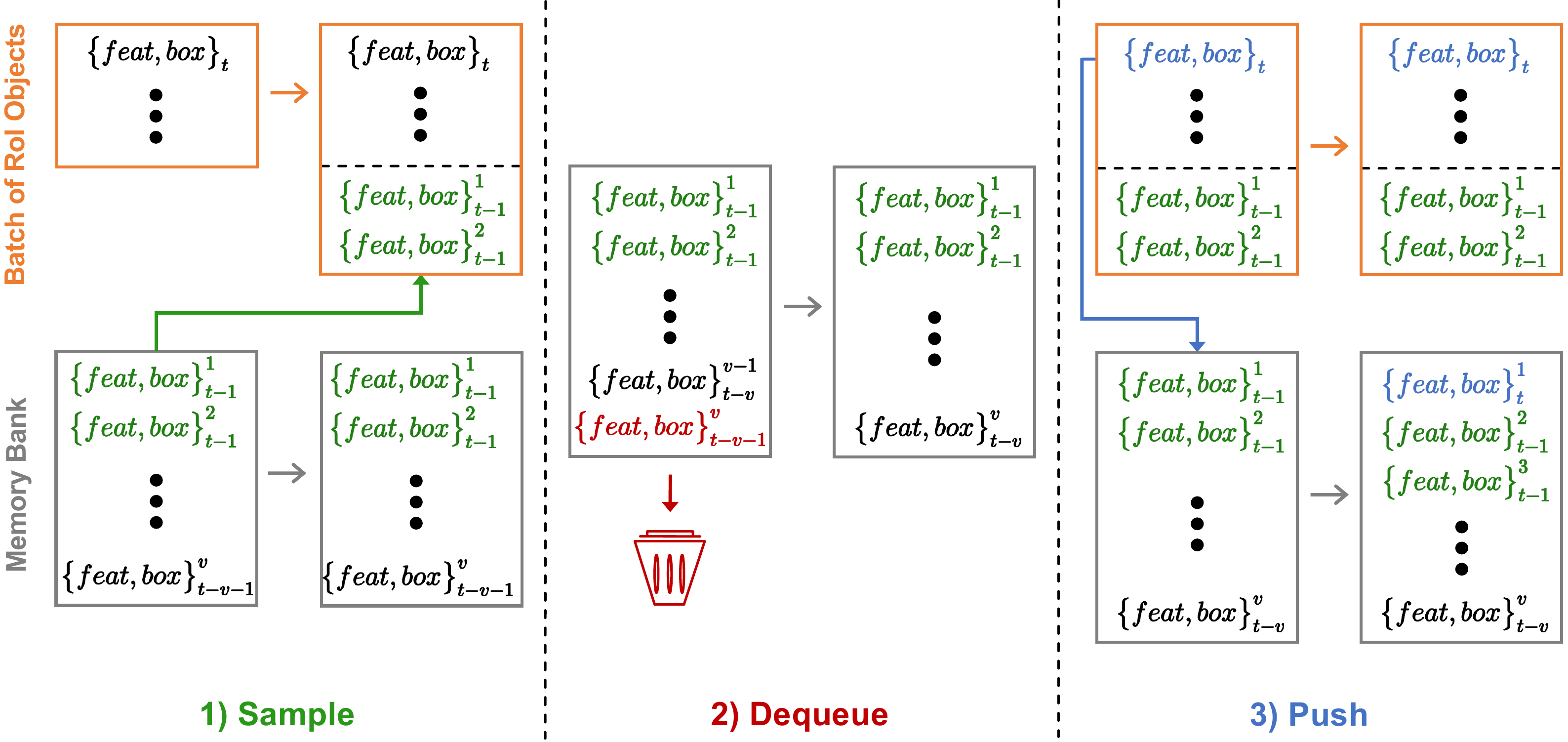}
\caption{Our memory bank three operations are shown here in the sequential order in which they will be used during training: 1) Sampling, 2) Dequeuing, 3) Pushing.}
\label{fig:ops}
\vspace{-0.4cm}
\end{figure}

\textbf{Setup.}
The memory bank \textit{M} consists of several independent queues for each targeted object class as seen in~\figref{fig:bank}. As we only want to repeat rare classes, the key classes in \textit{M} are only the rare classes. Note that the key classes can be set to any set of classes if desired. Thus, we denote each category queue in \textit{M} as $\bm{q}_r$, $r \in \bm{S}_r$. To improve efficiency and space, each $\bm{q}_r$ can only store a maximum amount of \textit{v} samples. We emphasize that our queues are not traditional queues in the sense that sampling from a queue does not remove the sample from the queue. Finally, we note that our memory bank is only utilized during training. Evaluation proceeds as normal without any augmentation. We describe the three main operations used on our memory bank and illustrate all operations in~\figref{fig:ops}.

\textbf{Pushing to Memory Bank.}
The memory bank consists of only object-level samples. In order to achieve that, \textit{M} is populated by RoI object features with class labels and their bounding box coordinates. Within our framework where we use Mask R-CNN as our architecture, we obtain RoI features and bounding boxes from the fully connected layer immediately prior to the classification and bounding box regression branches. However, we emphasize that any RoI level features can be used in the memory bank. We illustrate the location of our feature extraction in~\figref{fig:bank}. At training iteration \textit{t} with training batch \textit{B}, all objects proposed in image \textit{i} are denoted as $o^j_i$, $j \in (1, \dots ,k)$. If any object category $c^j_i \in \bm{S}_r$, we \texttt{push} the RoI features and bounding boxes, denoted as $\{feat, box\}_t$, into the top of \textit{M}'s category queue $\bm{q}_r$. We iterate through all images and objects to discover any RoI feature and bounding box to add to \textit{M}. 

\textbf{Dequeuing from Memory Bank.}
At any point when any queue $\bm{q}_r$ reaches its maximum space limit \textit{v}, we have to dequeue feature and proposal pairs from $\bm{q}_r$. We \texttt{dequeue} $\bm{q}_r$ from the bottom of the queue, where the oldest samples are located. As illustrated in~\figref{fig:ops}, when we want to push a pair $\{feat, box\}_t$ into a full $\bm{q}_r$, we first \texttt{dequeue} from $\bm{q}_r$ and then push $\{feat, box\}_t$ into $\bm{q}_r$.

\textbf{Sampling from Memory Bank.}
A batch cannot sample from a category \textit{r} from queue $\bm{q}_r$ in \textit{M} until $\bm{q}_r$ is populated with at least 1 sample. $\bm{q}_r$ is populated immediately after the first image containing $r$ is observed during training. Once $\bm{q}_r$ is populated, we can \texttt{sample} from $\bm{q}_r$ to augment any training batch. At training iteration \textit{t} with training batch \textit{B}, we augment the batch with object-level samples if necessary. For each $I_i$ in B, we augment the batch if there exist any $r \in \bm{S}_r$. Specifically, we augment the batch by sampling from $\bm{q}_r$ \textit{x} number of feature and proposal pairs $\{feat, box\}_{t-l}$, $l \in (1, \dots ,x)$. The number of samples \textit{x} can be changed as desired. From here, training proceeds as normal towards classification and bounding box regression. 

In summary, if there exists any target category in a current batch, the class' queue 1) \texttt{dequeues} if necessary, 2) \texttt{samples} additional features with ground truth classes and bounding boxes from the queue, and 3) \texttt{pushes} the current feature, ground truth class, and bounding box into the queue. We emphasize that through memory bank we are able to augment batches with object samples from different images, model snapshots, and image augmentations.

\begin{figure*}[ht!]
\centering
    \begin{subfigure}{0.24\textwidth}
      \centering
      \includegraphics[width=\textwidth]{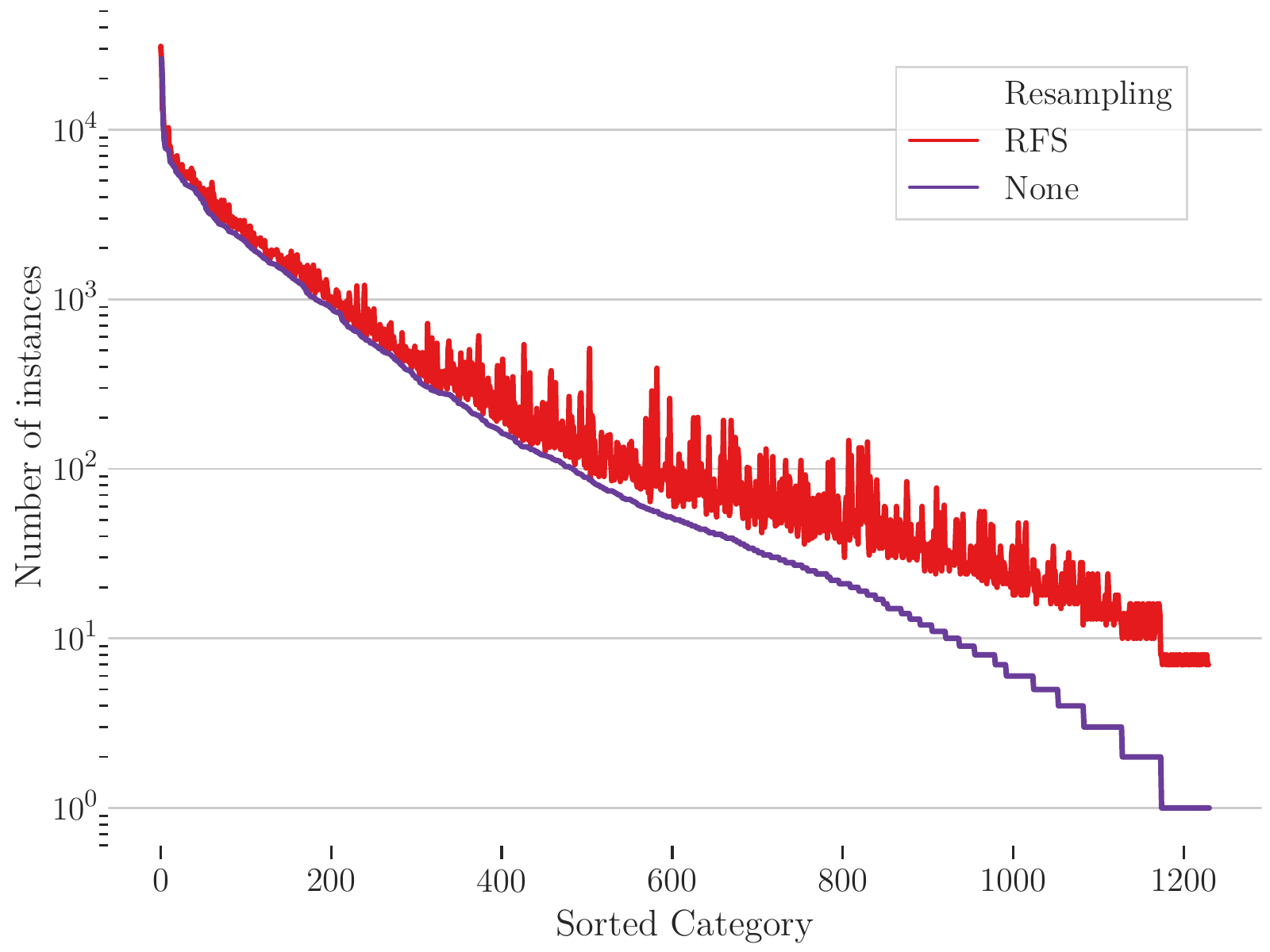}\vspace{-0.2cm}
      \subcaption{}
      \label{fig:rfs}
    \end{subfigure}
    \hfill
    \begin{subfigure}{0.24\textwidth}
      \centering
      \includegraphics[width=\textwidth]{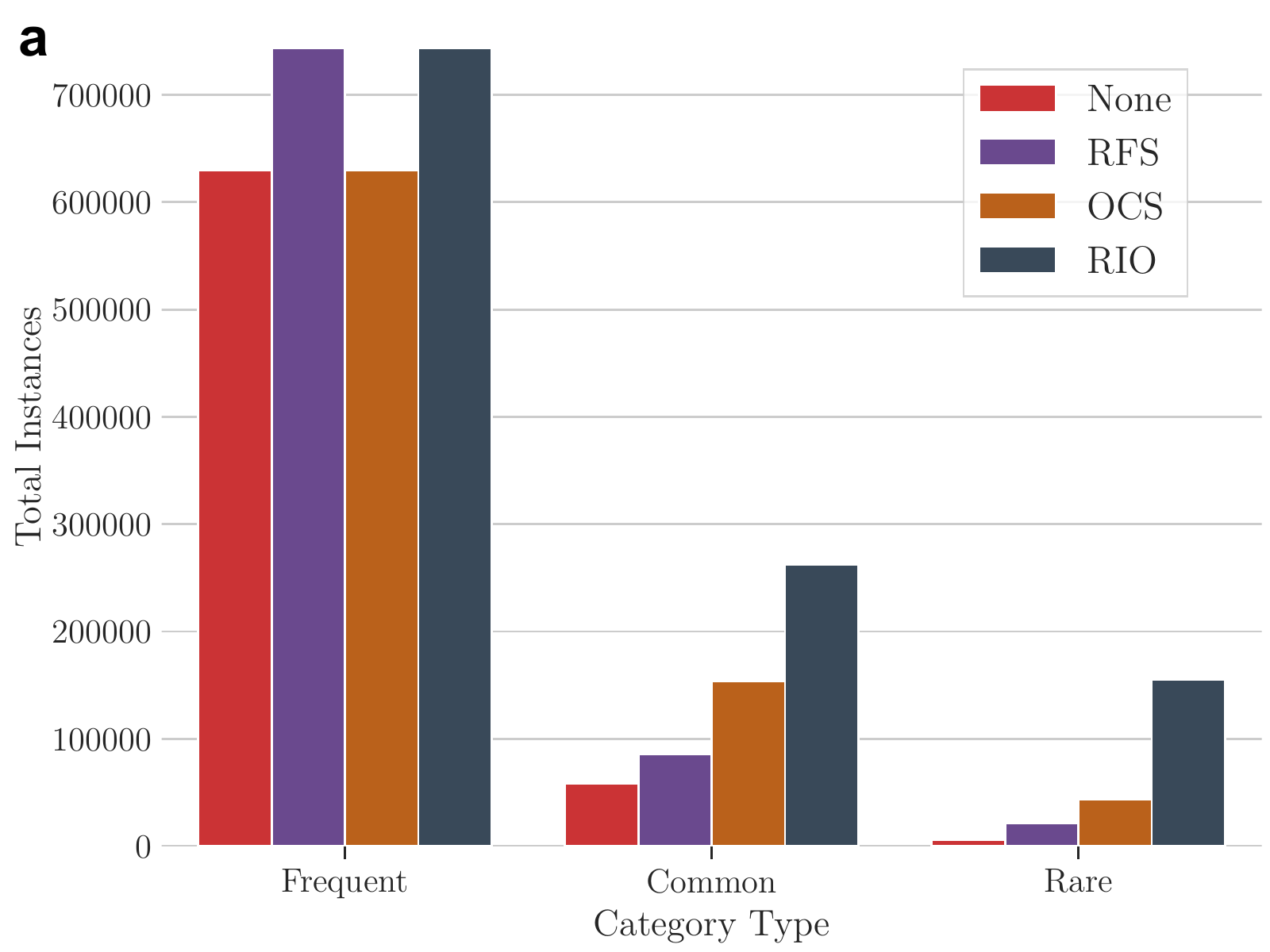}\vspace{-0.2cm}
      \subcaption{}
      \label{fig:total_insts}
    \end{subfigure}
    \hfill
    \begin{subfigure}{0.24\textwidth}
      \centering
      \includegraphics[width=\textwidth]{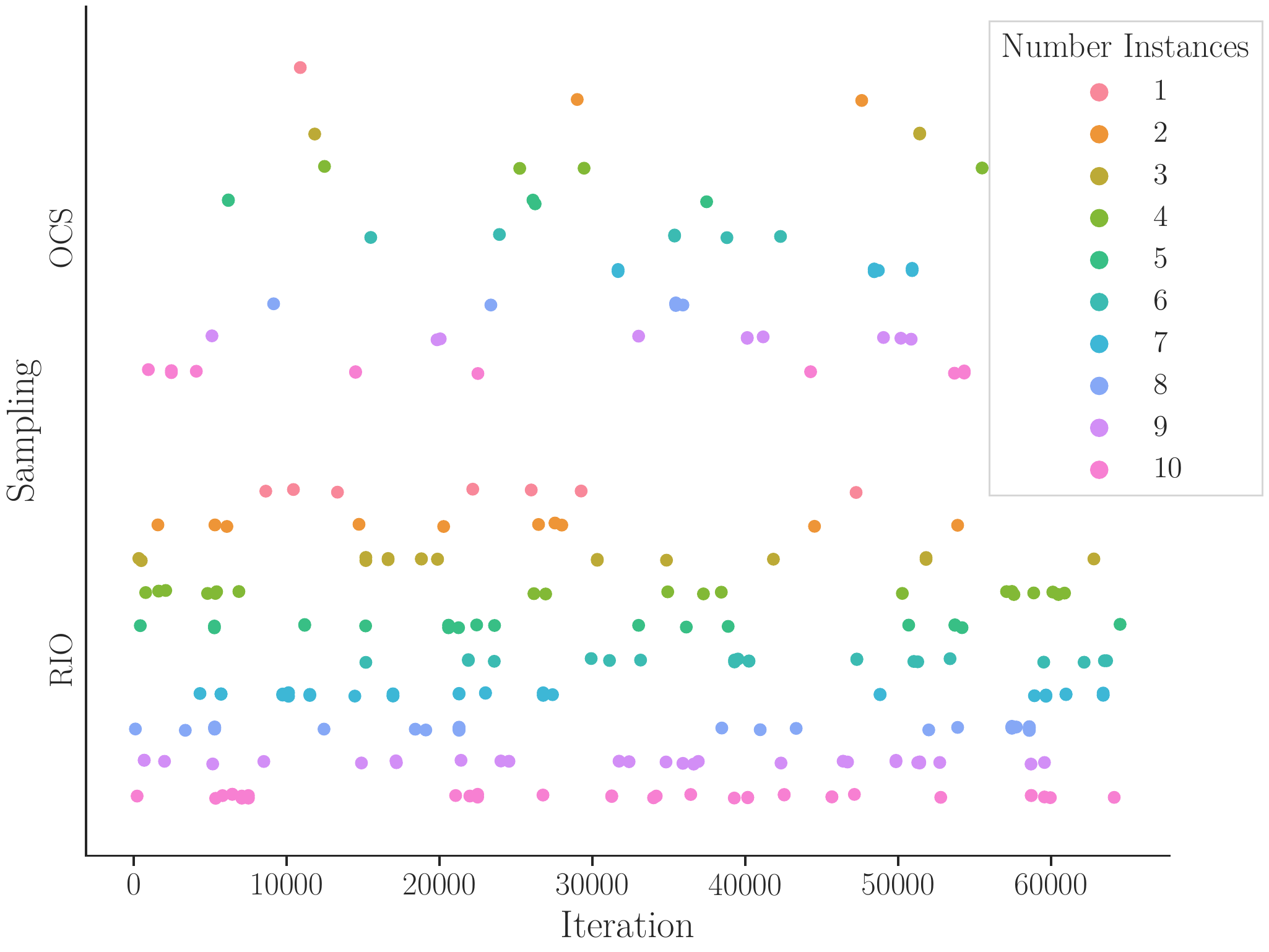}\vspace{-0.2cm}
      \subcaption{}
      \label{fig:bank_updates}
    \end{subfigure}
    \hfill
    \begin{subfigure}{0.24\textwidth}
      \centering
      \includegraphics[width=\textwidth]{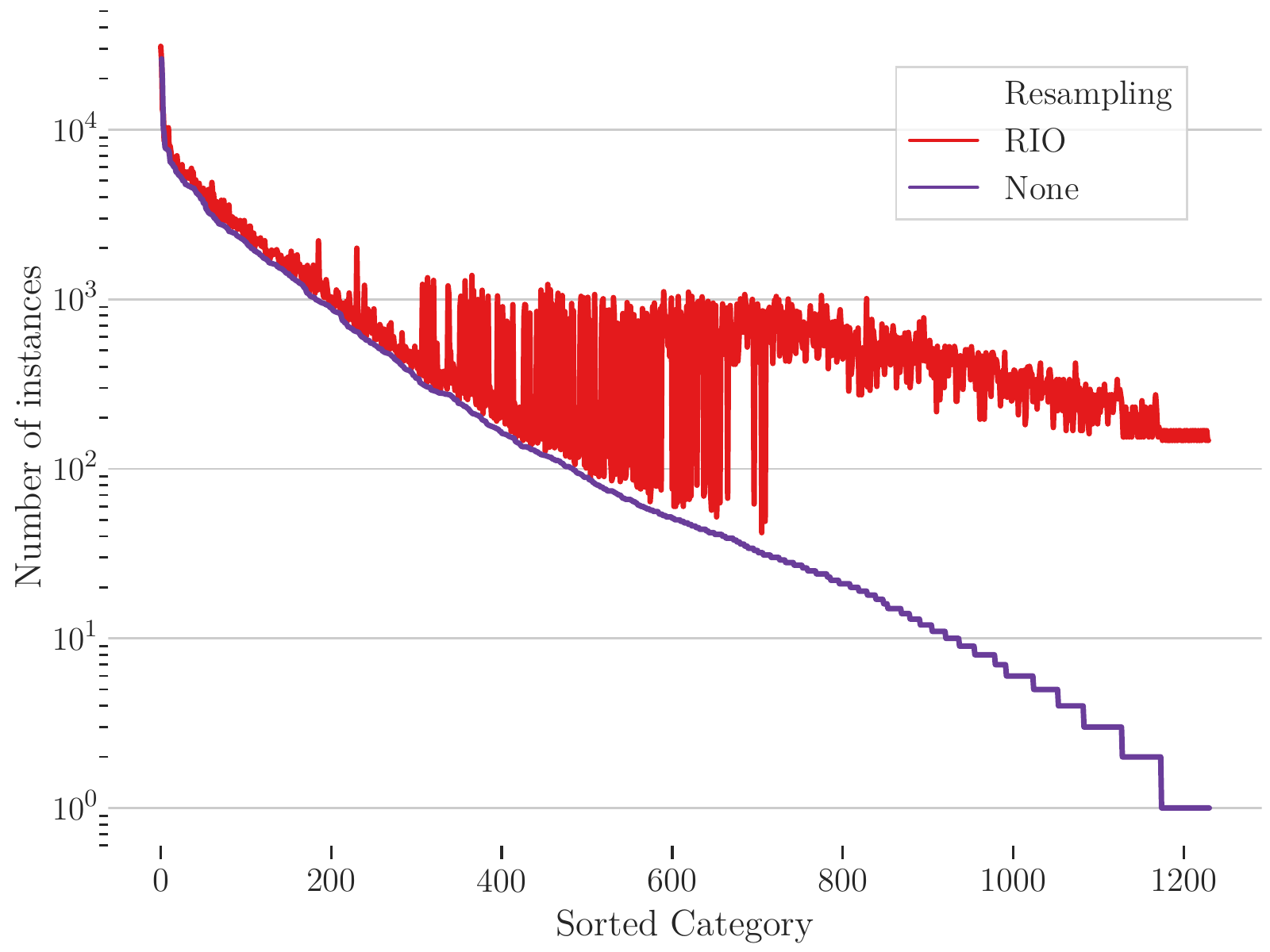}\vspace{-0.2cm}
      \subcaption{}
      \label{fig:rfs_ocs}
    \end{subfigure}
\vspace{-0.25cm}
\caption{Image-level and object-level resampling analysis. Start from the leftmost to the rightmost figure: (a) The number of instances per category with image resampling (RFS). (b) Total number of instances per category type. (c) Memory bank update frequency for some rare classes. (d) The number of instances per category with image and object resampling (RIO).}
\label{fig:analysis}
\vspace{-0.15cm}
\end{figure*}

\subsection{Resampling Strategy}
In this section, we explore different resampling strategies and analyze their effects in balancing the image/object-level distributions. We also consider their roles in designing our resampling policy with memory bank. 

\textbf{Dataset.}
We report our analysis on the LVIS version 0.5 dataset. LVIS version 0.5 contains 1230 classes, split into 454 rare categories, 461 common categories, and 315 frequent categories. LVIS defines classes with $> 100$ images as frequent, classes with $> 10$ but $< 100$ images as common, and classes with $\leq 10$ images as rare. 

\textbf{Image Resampling.}\label{sec:rfs}
Repeat factor sampling (RFS)~\cite{gupta2019lvis} has been a baseline image resampling method for LVIS. This method is not new, and is thus not part of our contribution. RFS is a pre-processing method that dictates which images are repeated per epoch. First, for each category \textit{c}, RFS computes the fraction of images that contain it and denotes it as $f(c)$. Next, for each category \textit{c}, it computes a category level ``repeat factor'' $r(c) = max(1, \sqrt{t/f(c)}$, where \textit{t} is a hyperparameter defaulted to 0.001. Finally, for each image \textit{I} containing \textit{k} unique categories, an image-level ``repeat factor'' is computed as $r(I) = max(1, r(c))$, $c \in 1, \dots , k$. Intuitively, the rarest category is used to compute the repeat factor for each image. 

Although RFS is an effective resampling strategy for detection, its image-based resampling has certain limitations for long-tailed detection problems. Images often contain a mixture of frequent, common, and rare categories. Thus, resampling an image causes all present objects to be resampled, not just the ones from rare classes. We illustrate this phenomenon by showing for an epoch the number of object instances per category in~\figref{fig:rfs}. One can observe that the frequent classes are also increased. Similarly, we show the total number of object instances for overall frequent, common, and rare categories in~\figref{fig:total_insts}. Frequent classes as a whole shows an 18\% relative increase in the number of object instances. Ideally, the focus of resampling in detection is to target sufficiently equalized object-level distributions. Yet manipulating object-level distribution is fundamentally difficult with an image-based resampling strategy. We therefore address this pitfall through object-level resampling. 

\textbf{Object Resampling.}
Object-centric sampling (OCS) specifically resamples targeted object instances as opposed to a whole image. We achieve this by augmenting a batch that has a targeted class with a set amount of additional features (default at 20). These features are stored in a memory bank that continuously updates when a targeted object is seen. For a given epoch, the object-level sampling effect on the number of object instances is shown in~\figref{fig:total_insts}. OCS is able to specifically resample classes with fewer instances, such as those in rare and a subset of common classes, without resampling additional frequent classes. However, OCS relies on an updating memory bank. The memory bank is only allowed to update when a batch contains an image with targeted objects. Thus, the success of OCS is restricted by the number of images per epoch. In~\figref{fig:bank_updates}, we can observe the frequency of memory bank updates for the most common types of rare classes that have 1-11 total object instances. In a worse case scenario without image resampling, a rare class seen in only 1 image can only update the memory bank once in an epoch. Further, object instances seen only in the beginning of an epoch lack more mature features found later in the epoch. The lack of updates reduces the amount of augmentation and better trained features over time. 

\textbf{Proposed Resampling Strategy.}
When considering a resampling policy for localization, we discover that neither image resampling nor object resampling is sufficient as a stand-alone strategy. Image resampling suffers from resampling redundant frequent classes, and object resampling is inherently restricted by the number of image appearances in an epoch that allows for memory bank updates. We hypothesize that localization resampling requires both image-level and object-level resampling, where each is complimentary to the other. As demonstrated in~\figref{fig:bank_updates}, image resampling allows the feature memory bank to update more frequently. Similarly, in~\figref{fig:total_insts}, the inclusion of object resampling increases the number of rare and common classes' instances without increasing the total frequent classes' instances. Conclusively, we propose RIO, a resampling strategy that jointly resamples at both image-level and object-level as a comprehensive resampling strategy for object detection. We illustrate RIO's effects on a training epoch by showing the number of object instances per category in~\figref{fig:rfs_ocs}.

%% file: sec4_exper.tex
\section{Experiments}
In this section we detail our experiments using RIO on the LVIS dataset. We demonstrate the effectiveness of our method against other resampling methods and previous top performing non-sampling methods on LVIS.

\subsection{Experimental Setup}
\textbf{Implementation.}
As detailed in \secref{sec:bank}, our memory bank targets specific classes to store. With LVIS v0.5, we target the infrequent categories with images $\leq 30$ samples. Thus, we focus on all LVIS' rare classes and a subportion of its common classes. Specifically, by using classes with $\leq 30$ samples, we target a total of 706 classes in our memory bank. Additionally, our experiments use Mask R-CNN with a ResNet-50 backbone and Feature Pyramid Network (FPN)~\cite{he2017mask}. Our networks are trained and tested with Detectron2~\cite{wu2019detectron2} on PyTorch version 1.5.1. Images are resized such that their shorter and longer edges are 800 and 1333 pixels, and are augmented with only horizontal flipping. Unless specified otherwise, models are trained with batchsize 16, base learning rate 0.02, and weight decay 0.0001 on 4 Tesla V100s with 32GB memory. Following standard Detectron2 settings, RPN uses only 256 regions for training. All models train for 90000 iterations, with decay at 60k and 80k iterations. During testing, 100 detections are allowed per image. We note that Mask R-CNN basic classifier layer is traditionally set as a linear classifier. However, several papers in long-tailed classification and detection~\cite{chen2019closer, wang2020frustratingly} have demonstrated the success of cosine layer for both classification and detection. Thus, our model uses cosine layer (Cos) as well. Lastly, our memory bank has a fixed maximum size \textit{v} in order to constrain the amount of memory taken. In all experiments we set \textit{v} to be 60 samples. Our models augment each batch by 20 samples for each targeted memory bank class that exists in the batch.

\textbf{Evaluation Metric.}
We follow the LVIS evaluation metrics and report the APs of the three category domains and denote them as follows: AP\textsubscript{f} for frequent classes, AP\textsubscript{c} for common classes, and AP\textsubscript{r} for rare classes. As we use Mask R-CNN, we report the overall box AP and mask AP of detection and segmentation as AP\textsubscript{b} and AP\textsubscript{m}, respectively. Lastly, we also report AP\textsubscript{50} and AP\textsubscript{75}, which refer to overall AP with $IoU=50$ and $IoU=75$ respectively. 

\subsection{Main Results on LVIS v0.5}
\textbf{Rebalancing Baselines.}
We first establish a few rebalancing baselines which feature simplicity using either resampling or reweighting. We compare against existing methods such as repeat factor sampling (RFS)~\cite{gupta2019lvis}, class-aware sampling~\cite{shen2016relay}, class-balanced loss~\cite{cui2019class}, and equalization loss (EQL)~\cite{tan2020equalization}. RFS was introduced along with LVIS, where each image is repeated if the rarest category in that image is below a certain threshold. The amount of repetition is determined by exactly how many images the rarest category contains as detailed in~\secref{sec:rfs}. Class-aware sampling first samples a category and then uniformly samples a random image that contains the category~\cite{shen2016relay}. Class-balanced loss weighs the loss by using the number of samples for each class to rebalance the loss~\cite{cui2019class}. Similar in nature to class-balanced loss, EQL's contribution is not explicitly image or object resampling. They ignore the gradient from samples of frequent categories for the rare categories. We report 2 sets of EQL performance. EQL's GitHub implementation notes that its results are better than the ones reported in the original paper because on top of horizontal flipping they include features such as scale jitter, class-specific mask head, and better ImageNet pretrained models which come along with Detectron2. In addition to the above rebalancing methods, we also report the performance of cosine layer baseline for ablation study.

\begin{table}[t]
\caption{Ablation study and comparison to resampling and reweighting baselines on LVIS v0.5, where $\dagger$ and $\ddag$ indicate results from the paper and GitHub of~\cite{tan2020equalization}, respectively.}
\label{tab:samps}
\vspace{0.1cm}
\resizebox{\linewidth}{!}{ 
\begin{tabular}{lcccccl}
\toprule
Detection & AP\textsubscript{b} & AP\textsubscript{50} & AP\textsubscript{75} & AP\textsubscript{r} & AP\textsubscript{c} & AP\textsubscript{f} \\
\midrule
Baseline (Cos) & 22.0 & 36.6 & 22.6 & 5.3 & 20.9 & \textbf{30.1} \\
Cls-Aware Smp.$\dag$ & 18.4 & - & - & - & - & - \\
Cls-Bal. Loss$\dag$ & 21.0 & - & - & - & - & - \\
EQL$\dag$ & 23.3 & - & - & - & - & - \\ 
EQL Github$\ddag$ & 23.6 & 38.3 & 25.2 & 8.5 & 23.9 & 29.3 \\
RFS (Cos) & 24.3 & 40.3 & 24.8 & 13.1 & 23.9 & 29.3 \\
\hdashline
OCS (Ours) & 24.4 & 40.0 & 25.4 & 10.4 & 24.4 & 29.9\\
RIO (Ours) & \textbf{25.7} &\textbf{41.8}& \textbf{26.7}& \textbf{17.2} & \textbf{25.1} & 29.8 \\
\specialrule{0.75pt}{2pt}{2pt}
Segmentation & AP\textsubscript{m} & AP\textsubscript{50}& AP\textsubscript{75} & AP\textsubscript{r} & AP\textsubscript{c} & AP\textsubscript{f} \\
\midrule
Baseline (Cos) & 22.9 & 35.4 & 24.2 & 6.7 & 23.1 & \textbf{29.0} \\
Cls-Aware Smp.$\dag$ & 18.5 & 31.1 & 18.9 & 7.3 & 19.3 & 21.9 \\
Cls-Bal. Loss$\dag$ & 20.9 & 33.8 & 22.2 & 8.2 & 21.2 & 25.7 \\ 
EQL$\dag$ & 22.8 & 36.0 & 24.4 & 11.3 & 24.7 & 25.1 \\
EQL Github$\ddag$ & 24.0 & 36.6 & 26.0 & 9.4 & 25.2 & 28.4 \\
RFS (Cos) & 25.1 & 38.9 & 26.6 & 15.1 & 25.4 & 28.6 \\
\hdashline
OCS (Ours) & 24.8 & 38.2 & 26.0 & 12.0 & 25.7 & 28.7 \\
RIO (Ours) & \textbf{26.0} &\textbf{39.7} & \textbf{28.0} & \textbf{18.9} & \textbf{26.2} & 28.5 \\
\bottomrule
\end{tabular}
}
\end{table}

\textbf{Object Detection.}
As shown in~\tabref{tab:samps}, our object-centric sampling (OCS) baseline achieves a 24.4\% overall box AP, which is slightly higher than RFS. Notably, OCS outperforms EQL GitHub with a 22.4\% relative improvement on rare AP. This increase illustrates the effectiveness of object resampling alone without additional image repeat. Unified resampling using RIO further achieves 25.7\% overall AP, a 5.8\% relative improvement from RFS alone. In addition, RIO achieves the best performance on rare categories with a 17.2\% AP, which is a 31.3\% relative improvement from RFS and a 102.3\% relative improvement from EQL GitHub.

\textbf{Instance Segmentation.} As shown in~\tabref{tab:samps}, RIO achieves a 26.0\% overall mask AP, which is 0.9\% better than RFS, 3.2\% better than EQL, 2.0\% better than EQL GitHub, 6.0\% better than class-balanced loss, and 7.5\% better than class-aware sampling. OCS alone outperforms EQL whereas RIO further improves both rare and common categories.

\textbf{State-of-the-arts.}
Besides the above rebalancing baselines, we also compare with state-of-the-art methods in~\tabref{tab:sota_all}, and illustrate the relative improvement over them in~\tabref{tab:relativesota}. We consider several top performing methods: 1) RFS + Fine-Tuning (FT)~\cite{wang2020frustratingly}, 2) RFS + EQL~\cite{gupta2019lvis}, 3) Forest R-CNN~\cite{wu2020forest}, 4) Balanced Group Softmax (BAGS)~\cite{li2020overcoming}. Note that Forest R-CNN is a strong but relatively complicated framework which requires additional class hierarchy structure and supervision.

\textbf{Object Detection.}
Since RFS + FT only provides detection results, we compare to it only on this task. Our method outperforms by 1.3\% on overall box AP, 1.1\% on rare AP, 0.8\% on common AP, and 2.1\% on frequent AP. Our method achieve a relative 5.2\% improvement on the overall box AP and a relative 1.9\% improvement on the rare AP. Our rare AP is 0.3\% higher than Forest R-CNN, 1.2\% higher than RFS + EQL, and 2.2\% higher than BAGS, which correspond to 1.9\%, 7.4\%, and 14.6\% relative improvements, respectively. We note that BAGS mainly targets the overall performance and emphasizes less on the rare classes. Although Forest R-CNN is less imbalanced, its performance on rare classes still has space to improve. On the other hand, RIO achieves state-of-the-art rare AP, while maintaining competitive overall AP. This provides a good trade-off between rare and overall performance and better ensures model fairness.

\textbf{Instance Segmentation.}
Our improvement on rare AP remains significant in instance segmentation. Our method outperforms Forest R-CNN by a relative 1.5\% on overall mask AP, 3.1\% on rare AP, and 3.3\% on frequent AP. Compared to BAGS and RFS + EQL, our rare AP is 0.9\% and 1.7\% higher, or 5.0\% and 9.7\% relative improvements.

\begin{table}[t]
\caption{Comparison to state-of-the-art methods on LVIS v0.5 with a ResNet-50 backbone. $\dag$ indicates results from the GitHub.}
\label{tab:sota_all}
\vspace{0.1cm}
\centering
\resizebox{\linewidth}{!}{ 
\begin{tabular}{lcccccc}
\toprule
Detection & AP\textsubscript{b} & AP\textsubscript{50} & AP\textsubscript{75} & AP\textsubscript{r} & AP\textsubscript{c} & AP\textsubscript{f} \\
\midrule
RFS + FT & 24.4 & 40.0 & 26.1 & 16.9 & 24.3 & 27.7 \\
RFS + EQL$\dag$ & 25.4 & 41.1 & 27.0 & 16.0 & 25.4 & 29.1 \\
Forest R-CNN & 25.9 &\textbf{42.7} & 27.2 & 16.9 & 26.1 & 29.2 \\
BAGS & 25.8 & - & - & 15.0 & 25.5 & \textbf{30.4} \\
\hdashline
RIO (Ours) & 25.7 & 41.8 & 26.7 & 17.2 & 25.1 & 29.8 \\
RIO + EQL (Ours) & \textbf{26.1} & 41.9 & \textbf{27.3} & \textbf{18.6} & \textbf{26.2} & 29.0 \\
\specialrule{0.75pt}{2pt}{2pt}
Segmentation & AP\textsubscript{m}  & AP\textsubscript{50} & AP\textsubscript{75} & AP\textsubscript{r} & AP\textsubscript{c} & AP\textsubscript{f} \\
\midrule
RFS + EQL$\dag$ & 26.1 & 39.6 & 27.6 & 17.2 & 27.3 & 28.2 \\
Forest R-CNN & 25.6 & \textbf{40.3} & 27.1 & 18.3 & 26.4 & 27.6 \\
BAGS & 26.2 & - & - & 18.0 & 26.9 & \textbf{28.7} \\
\hdashline
RIO (Ours) & 26.0 & 39.7 & \textbf{28.0} & \textbf{18.9} & 26.2 & 28.5 \\
RIO + EQL (Ours) & \textbf{26.3}& 39.8 & 27.9 & 18.8 & \textbf{27.4} & 27.9 \\
\bottomrule
\end{tabular}
}
\end{table}

\begin{table}[!t]
\caption{Relative performance gain (\%) of our methods compared to other methods in~\tabref{tab:sota_all}. $\dag$ indicates results from the GitHub.}
\label{tab:relativesota}
\vspace{0.1cm}
\centering
\resizebox{\linewidth}{!}{ 
\addtolength{\tabcolsep}{-2pt} 
\begin{tabular}{llcccccl}
\toprule
\multirow{2.3}*{Method} & \multirow{2.3}*{Comp Method} & \multicolumn{3}{c}{Detection} & \multicolumn{3}{c}{Segmentation}
\\\cmidrule(lr){3-5}\cmidrule(lr){6-8}
           && AP\textsubscript{b}  & AP\textsubscript{r} & AP\textsubscript{c}    & AP\textsubscript{m}  & AP\textsubscript{r} & AP\textsubscript{c}\\
\midrule
RIO & RFS + FT & \textbf{5.2} & 1.9 & \textbf{3.2} & - & - & - \\
RIO & RFS + EQL$\dag$ & 1.0 & 7.4 & -1.4 & -0.5 & \textbf{9.7} & -4.0 \\
RIO & Forest R-CNN & -0.9 & 1.9 & -3.9 & \textbf{1.5} & 3.1 & -0.7 \\
RIO & BAGS & -0.4 & \textbf{14.6} & -1.5 & -1.0 & 5.0 & -2.6 \\
\midrule
RIO + EQL & RFS + FT & \textbf{7.0} & 10.2 &\textbf{7.8} & - & - & - \\
RIO + EQL & RFS + EQL$\dag$ & 2.8 & 16.1 & 3.1 & 0.7 &\textbf{9.5} & 0.0 \\
RIO + EQL & Forest R-CNN & 0.8 &10.2 &0.4 &2.7 &2.9 &\textbf{3.7}\\
RIO + EQL & BAGS & 1.3 & \textbf{23.9} & 3.0 & 0.1 & 4.8 & 1.7\\
\bottomrule
\end{tabular}
}
\end{table}

\begin{table}[t]
\caption{Comparison to state-of-the-art methods on LVIS v0.5 with ResNet-101/ResNeXt-101. $\dag$ indicates results from the GitHub.}
\label{tab:arch}
\vspace{0.1cm}
\centering
\resizebox{\linewidth}{!}{ 
\begin{tabular}{lccccccc}
\toprule
Detection & Bkb & AP\textsubscript{b} & AP\textsubscript{50} & AP\textsubscript{75} & AP\textsubscript{r} & AP\textsubscript{c} & AP\textsubscript{f} \\
\midrule
RFS + FT & R101 & 26.2 & 41.8 & 27.5 & 17.3 & 26.4 & 29.6 \\
RFS + EQL$\dag$ & R101 & 27.1 & 43.0 & \textbf{29.1} & 15.9 & \textbf{27.9} & 30.6 \\
Forest R-CNN & R101 & 27.5 & \textbf{44.9} & 29.0 & \textbf{20.0} & 27.5 & 30.4 \\
\hdashline
RIO & R101 & 27.3 & 43.7 & 29.0 & 19.1 & 26.8 & \textbf{31.2} \\
RIO + EQL & R101 & \textbf{27.6} & 43.5 & 28.9 & 19.2 & 27.7 & 30.9 \\
\midrule
Forest R-CNN & X101 & \textbf{28.8} & \textbf{46.3} & \textbf{30.9} & \textbf{20.6} & \textbf{29.2} & 31.7\\
BAGS & X101 & 27.8 & - & - & 18.8 & 27.3 & 32.1\\
\hdashline
RIO & X101 & 28.6 & 45.2 & 30.5 & 19.0 & 28.0 & \textbf{33.0}\\
\specialrule{0.75pt}{2pt}{2pt}
Segmentation & Bkb & AP\textsubscript{m} & AP\textsubscript{50} & AP\textsubscript{75} & AP\textsubscript{r} & AP\textsubscript{c} & AP\textsubscript{f} \\
\midrule
RFS + EQL$\dag$ & R101 & 27.4 & 41.5 & 29.4 & 17.3 & \textbf{29.0} & 29.4 \\
Forest R-CNN & R101 & 26.9 & 42.2 & 28.4 & \textbf{20.1} & 27.9 & 28.3 \\
\hdashline
RIO & R101 & \textbf{27.7} & \textbf{42.3} & 29.0 & \textbf{20.1} & 28.3 & \textbf{30.0}\\
RIO + EQL & R101 & 27.6 & 41.6 & \textbf{29.6} & 19.8 & 28.6 & 29.6 \\
\midrule
Forest R-CNN & X101 & 28.5 & 43.8 & \textbf{30.9} & \textbf{21.6} & \textbf{29.7} & 29.7\\
\hdashline
RIO & X101 & \textbf{28.9} & \textbf{44.0} & \textbf{30.9} & 19.5 & \textbf{29.7} & \textbf{31.6}\\
\bottomrule
\end{tabular}
}
\end{table}

\textbf{EQL + RIO.}
We demonstrate that RIO can be complementary to other methods by combining with EQL. In~\tabref{tab:sota_all}, RIO + EQL achieves 26.1\% AP\textsubscript{b} and 26.3\% AP\textsubscript{m}, a state-of-the-art performance over all comparing methods. In~\tabref{tab:relativesota}, our relative improvements over comparing methods are comprehensive. We note that the improvements on rare box AP are significant, ranging from 10.2\% to 23.9\%. The promising results show that the benefits from our results remain even when utilized with another competitive method, which validates the value of RIO as a useful approach.

\textbf{Generalization across Backbones.}
We present additional results trained with ResNet-101 and ResNeXt-101 (32$\times$8d) backbones in~\tabref{tab:arch}. We show that the improvements of RIO and RIO + EQL are consistent across different network architectures. In fact, the improvements become more significant as RIO alone comprehensively outperforms RFS + EQL and BAGS. RIO is comparable with Forest R-CNN on detection and slightly better on instance segmentation.

\begin{figure*}[th]
\begin{center}
\includegraphics[width=\linewidth]{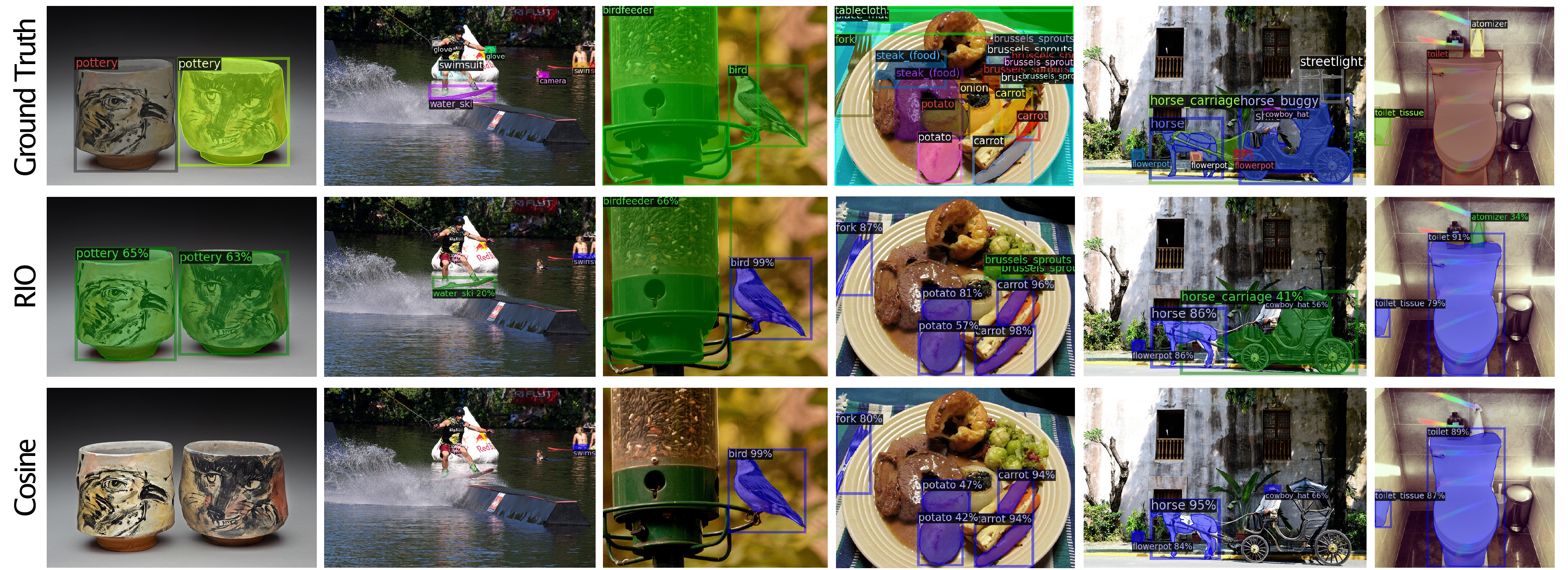}
\end{center}
\vspace{-0.2cm}
\caption{Detection and instance segmentation visualizations of different models. The targeted rarer categories are colored in green and the frequent classes are colored in blue. Note that certain object categories such as human are not part of the LVIS annotations. Thus these object categories are not detected.}
\label{fig:visualization}
\end{figure*}

\begin{figure}[t]
\centering
    \begin{subfigure}{0.238\textwidth}
      \centering
      \includegraphics[width=\textwidth]{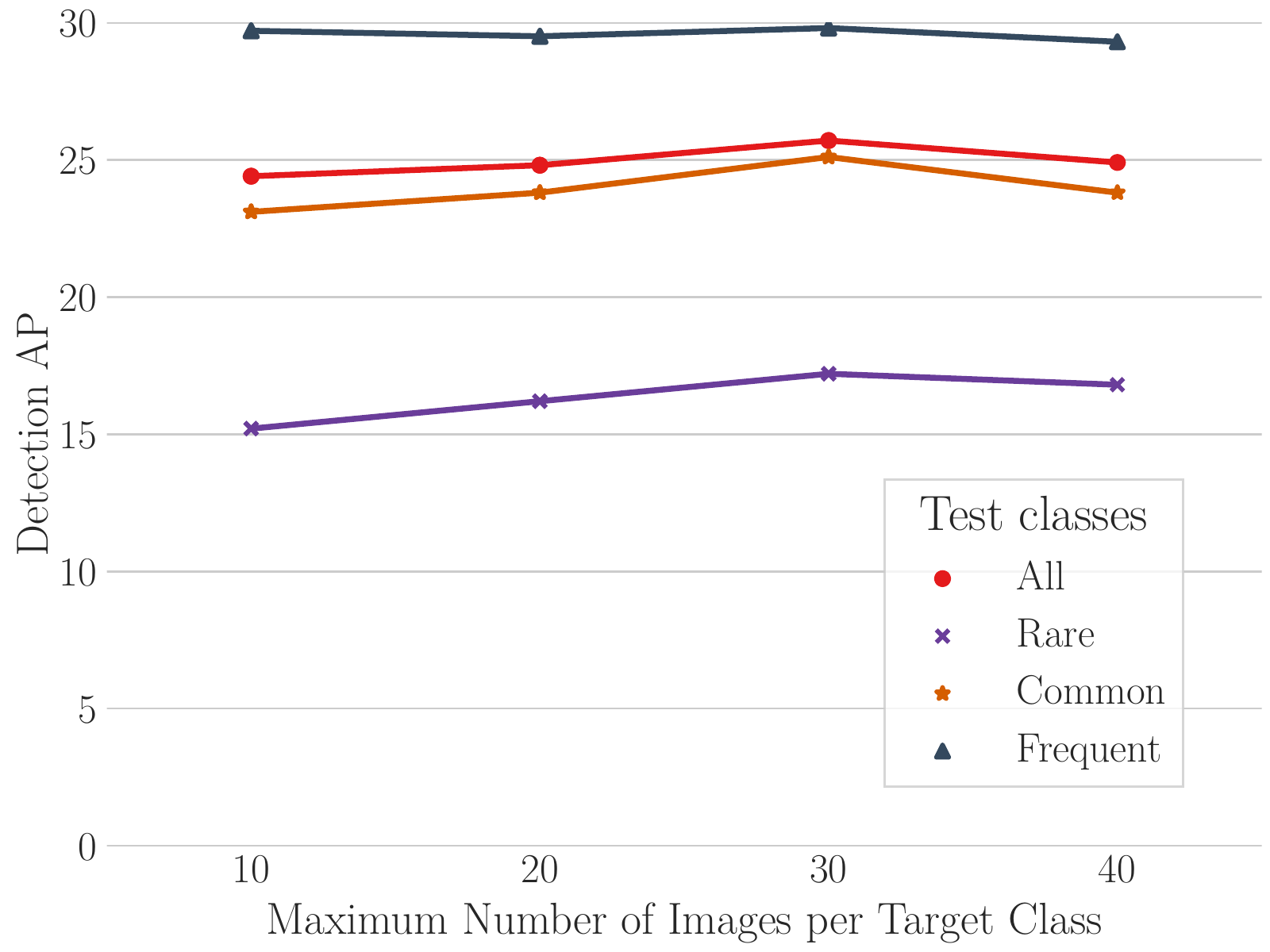}\vspace{-0.2cm}
      \subcaption{}
      \label{fig:fig6a}
    \end{subfigure}
    \hfill
    \begin{subfigure}{0.238\textwidth}
      \centering
      \includegraphics[width=\textwidth]{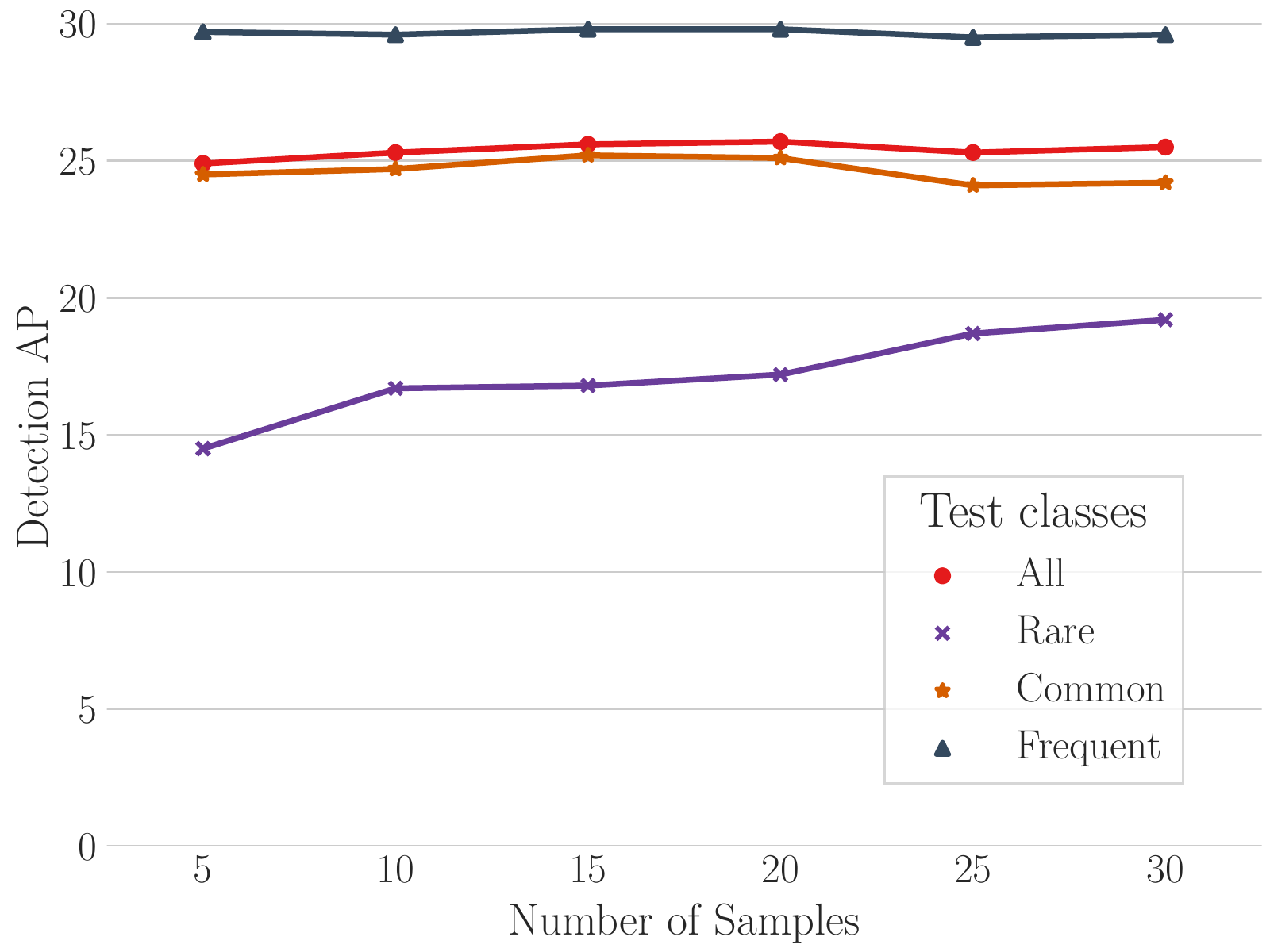}\vspace{-0.2cm}
      \subcaption{}
      \label{fig:fig6b}
    \end{subfigure}
    \hfill
    \begin{subfigure}{0.238\textwidth}
      \centering
      \includegraphics[width=\textwidth]{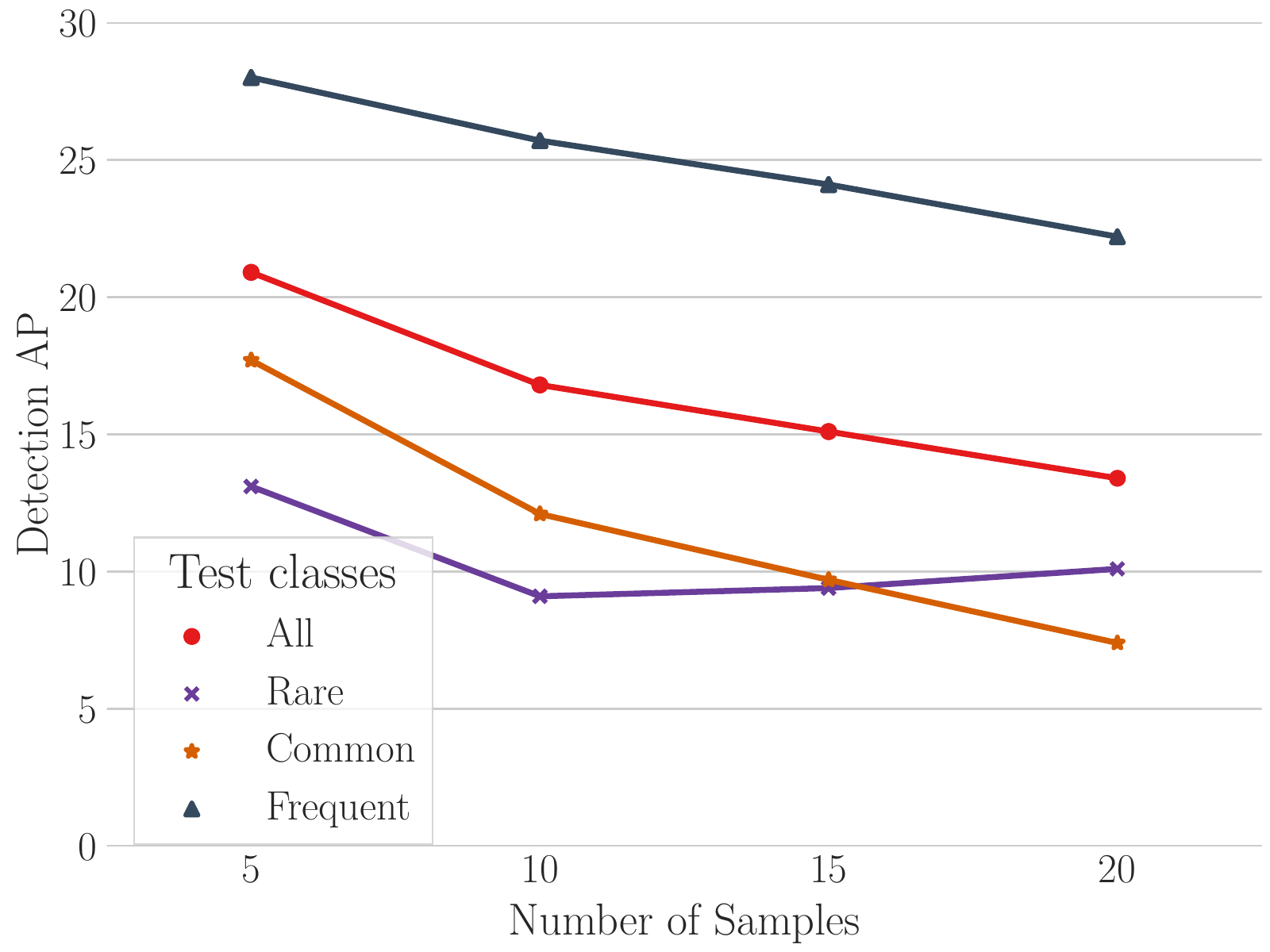}\vspace{-0.2cm}
      \subcaption{}
      \label{fig:fig6c}
    \end{subfigure}
\vspace{-0.1cm}
\caption{Ablations studies. (a) The effect of targeted classes on detection performance. (b) The effect of sampled object number from a memory bank on detection performance. (c) Baseline method which simply repeats the RoI features of targeted classes in each batch instead of using memory bank.}
\label{fig:ablations}
\vspace{-0.2cm}
\end{figure}

\textbf{Analysis on Targeted Class.}
We study the effect of targeted classes with memory bank. A class is targeted if the number of images containing that class is less than a threshold. In this study, we vary the threshold as 10/20/30/40 and show the detection performance in~\figref{fig:fig6a}. Our study shows that repeating samples for classes that already have sufficient data decreases the performance. Alternatively, repeating samples for a more limited set of classes is also less effective. From our ablation we observe a peak performance when classes with less than 30 samples are targeted. 

\textbf{Analysis on Sample Number.}
We further conduct analysis on the number of objects retrieved each time from a memory bank. We vary this number from 5 to 30 and show the object detection performance in~\figref{fig:fig6b}. We observe that the best overall performance is achieved at 20 samples. Rare AP continue to increase as we increase this number, showing the significant benefit of memory replay on rare classes. However, the performance of more frequent classes tend to decrease mildly after the number exceeds 20, whereas a lower number does not contribute enough on rare AP.

\textbf{Analysis on Memory Replay Efficacy.}
We present a baseline which simply repeats the current rare objects' RoI features. There is no memory bank to sample additional samples from. \figref{fig:fig6c} shows that such strategy leads to a significant decrease in performance. We notice a consistent overall decrease in all classes except rare classes, even though reducing the repeat number tends to alleviate this issue. The study shows the efficacy of memory replay as it provides valuable feature augmentation from model update and object diversity, which is key to the improved performance.

\textbf{Additional Visualizations.}
We additionally visualize the detection and segmentation results in Fig.~\ref{fig:visualization}. One can see that RIO can improve the detection of rare class objects with significantly varying sizes, such as birdfeeder (large), horse carriage (medium) and atomizer (small).

\subsection{Main Results on LVIS v1.0}
We showcase our method on LVIS v1.0. Results can be seen in~\tabref{tab:lvis1}. Similar to LVIS v0.5, RIO makes significant gains in rare, common, overall classes with negligible effects on frequent classes. In detection and segmentation, RIO improves over RFS in overall performance by a relative 3.6\% and 3.4\% respectively. Importantly, within rare classes we see a significant relative improvement of 36.4\% and 29.9\% in detection and segmentation respectively. Furthermore, we show that RIO is compatible with another backbone, ResNeXt-101 (32$\times$8d) on LVIS v1.0 and observes consistent healthy improvements. Lastly, we would like to acknowledge that there are several top performing methods in the LVIS Challenge. However, all methods presented use several methods together, such as a much larger architecture, additional dataset annotations, and fine-tuning. 2020's LVIS winner combined more than 10 techniques.

\begin{table}[h!]
\vspace{-0.1cm}
\caption{Detection and segmentation results on LVIS v1.0.}
\label{tab:lvis1}
\centering
\vspace{0.1cm}
\resizebox{\linewidth}{!}{ 
\begin{tabular}{lccccccc}
\toprule
Detection &Bkb  & AP\textsubscript{b} & AP\textsubscript{50} & AP\textsubscript{75} & AP\textsubscript{r} & AP\textsubscript{c} & AP\textsubscript{f} \\
\midrule
Baseline (Cos) & R50 & 19.9 & 32.5 & 20.7	& 2.6 & 16.6 & \textbf{31.0} \\
RFS (Cos) & R50 & 23.3 & 37.7 & 24.5 & 10.7 & 21.6 & 30.6 \\
\hdashline
RIO & R50 & \textbf{24.1} & \textbf{38.6} & \textbf{25.4} & \textbf{14.6} & \textbf{22.3} & 30.4 \\
\midrule
Baseline (Cos) & X101 & 24.3 & 38.7 & 25.4 & 4.3 & 22.3 & \textbf{35.4} \\
RFS (Cos) & X101 & 27.7 & 43.0 & 29.7 & 14.8 & 26.6 & 34.5 \\
\hdashline
RIO & X101 & \textbf{28.5} & \textbf{44.0} & \textbf{30.6} & \textbf{18.6} & \textbf{26.9} & 34.6 \\
\specialrule{0.75pt}{2pt}{2pt}
Segmentation &Bkb & AP\textsubscript{m}  & AP\textsubscript{50} & AP\textsubscript{75} & AP\textsubscript{r} & AP\textsubscript{c} & AP\textsubscript{f} \\
\midrule
Baseline (Cos) & R50 & 19.5 & 30.7 & 20.4	& 2.9 & 17.3 & \textbf{29.2} \\
RFS (Cos) & R50 & 22.9 & 35.5 & 24.1 & 11.7 & 21.9 & 28.9 \\
\hdashline
RIO & R50 & \textbf{23.7} & \textbf{36.6} & \textbf{25.3} & \textbf{15.2} & \textbf{22.5} & 28.8 \\ 
\midrule
Baseline (Cos) & X101 & 23.8 & 36.6 & 25.1 & 5.4 & 22.7 & \textbf{33.0} \\
RFS (Cos) & X101 & 26.9 & 40.8 & 28.7 & 15.4 & 26.5 & 32.3 \\
\hdashline
RIO & X101 & \textbf{27.5} & \textbf{41.6} & \textbf{29.5} & \textbf{18.8} & \textbf{26.7} & 32.3 \\
\bottomrule
\end{tabular}
}
\end{table}

%% file: sec5_concl.tex
\section{Conclusion}
We proposed a novel object-centric memory replay framework which allows us to consider a joint resampling strategy at both image and object level (RIO) for object detection. As part of RIO, we proposed OCS as the first RoI-level sampling method introduced for multi-object tasks with implicit feature augmentation from the memory replay. RIO achieves the state-of-the-art performance on rare categories in LVIS while maintaining overall AP. Importantly, by emphasizing model performance on less represented categories without sacrifice to overall model performance, our method creates a more fair model for long-tailed localization.

%% file: sec6_ack.tex
\section*{Acknowledgment}
We would like to sincerely thank Achal Dave, Kenneth Marino, Senthil Purushwalkam and other NVIDIA colleagues for the discussion and constructive suggestions.